\documentclass{jpsj3}
\usepackage{txfonts}
\usepackage{color}
\usepackage{graphicx}
\usepackage{amsmath}
\usepackage{amsfonts}
\title{$L_1$-regularized Boltzmann machine learning\\ using majorizer minimization}

\author{Masayuki Ohzeki\thanks{mohzeki@i.kyoto-u.ac.jp}}
\inst{Department of Systems Science, Graduate School of Informatics, Kyoto University, 36-1 Yoshida Hon-machi, Sakyo-ku, Kyoto, 606-8501, Japan} 

\abst{
We propose an inference method to estimate sparse interactions and biases according to Boltzmann machine learning.
The basis of this method is $L_1$ regularization, which is often used in compressed sensing, a technique for reconstructing sparse input signals from undersampled outputs.
$L_1$ regularization impedes the simple application of the gradient method, which optimizes the cost function that leads to accurate estimations, owing to the cost function's lack of smoothness.
In this study, we utilize the majorizer minimization method, which is a well-known technique implemented in optimization problems, to avoid the non-smoothness of the cost function.
By using the majorizer minimization method, we elucidate essentially relevant biases and interactions from given data with seemingly strongly-correlated components.
}

\begin{document}

\maketitle

\section{Introduction}
Because massive amounts of structured and unstructured data continue to accumulate, the importance of effective big data analysis is rapidly increasing.
One well-known big data analysis tool is Boltzmann machine learning. This technique is physics-friendly, because it is a form of probability density defined by the Hamiltonian of the Ising model \cite{Ackley1985}.
We assume that the generative model has a bias on each variable, the magnetic field, and the pair-wise interactions between the different variables (i.e., the interaction between adjacent spins).
Boltzmann machine learning has proven effective, and has stimulated increasing interest in deep learning \cite{Hinton2006,Hinton2006sci,RG2014,Ohzeki2015}.
Deep learning typically needs large volumes of data for its implementation.
Currently, this demand is often satisfied because we are in the so-called big data era; however, we require hard computation as a return.
Thus, the study of Boltzmann machine learning may involve constructing a good approximation \cite{Sessak2009,Cocco2011,Cocco2012,Ricci2012,Yasuda2013,Raymond2013,Ohzeki2013}.
Otherwise, we require a novel method to achieve efficient learning, even from a small amount of data.

Effective big data analysis can produce a substantial amount of valuable information.
An objective of this analysis is to elucidate a small number of relevant quantities to describe the acquired data, a process known as variable selection.
The goal of data-driven science is to capture an essential portion of the generative model and to identify the characteristics that describe its origin.
In order to achieve this goal, sparseness may be imposed on the bias or pair-wise interactions of the generative model.
One successful approach is to employ the regularization of the $L_1$ norm of the bias and pair-wise interactions.
However, because of the $L_1$ norm's lack of differentiability, the application of the simple gradient method is not straightforward.
A different method employs a greedy algorithm, which seeks a small number of non-zero components satisfying some criteria.
Under some conditions, greedy algorithms can overcome the $L_1$ regularization \cite{Aurelien2014,Yamanaka2015}.
However, greedy algorithms depend on the properties of the parameters to be estimated; moreover, $L_1$ regularization cannot be discarded, because it has a wide range of applications and enables us to perform robust inference for various models.

In this study, we resolve the lack of smoothness by implementing a technique for $L_1$ regularization (often used in optimization studies), namely majorizer minimization \cite{Beck2009book,Beck2009}.
The technique reduces a "many-body" interaction problem to a "one-body" problem by introducing the majorizer of the original optimization problem with $L_1$ regularization.
This is a type of mean-field analysis used in statistical mechanics.
We must emphasize that this method does not change the optimal solution, and thus yields the exact optimal point under several optimized cost function conditions.

The remaining sections of the paper are organized as follows.
In the second section, we briefly review Boltzmann machine learning and the recent developments in this area.
In the third section, we introduce majorizer minimization, and obtain the algorithm to resolve the Boltzmann machine learning optimization problem, using $L_1$ regularization.
In the fourth section, we test our method with numerical experiments.
In the last section, we summarize our study.

\section{Boltzmann machine learning}
We assume that the generative model of the data ${\bf x} \in \{-1,1\}^N$ takes the form of the Ising model as
\begin{equation}
P({\bf x}|J,{\bf h}) = \frac{1}{Z(J,{\bf h})} \exp\left( \sum_{i=1}^N \sum_{j \in \partial i} J_{ij} x_i x_j + \sum_{i=1}^N h_ix_i \right),\label{GB}
\end{equation}
where $J_{ij}$ is a pair-wise interaction, $h_i$ is a bias, and $Z(J,{\bf h})$ is the partition function.
The sets of $J_{ij}$ and $h_i$ are denoted as $J$ and ${\bf h}$.
The number of components is represented by $N$.
The summation $j \in \partial i$ is calculated by summing the adjacent components to one denoted by $i$.
Boltzmann machine learning is used to estimate $J_{ij}$ and $h_i$ from snapshots of spin configurations, namely the given data, ${\bf x}^{(k)}$ for $k=1,2,\cdots,D$ by use of the Gibbs-Boltzmann distribution of the Ising model as in Eq. (\ref{GB}).
The standard method to estimate the parameters $J$ and ${\bf h}$ is the maximum-likelihood estimation \cite{Bishop2006} as
\begin{equation}
\left\{ J, {\bf h} \right\} = \arg \max_{J,{\bf h}}\left\{ \sum_{k=1}^D \log P({\bf x}^{(k)}|J,{\bf h})\right\}.
\end{equation}
In other words, we minimize the KL divergence between the generative model's distribution and the empirical distribution of the given data defined as
\begin{equation}
P_{\mathcal{D}}({\bf x}) = \frac{1}{D} \sum_{k=1}^D \delta({\bf x}- {\bf x}^{(k)}).
\end{equation}
The minimization of KL divergence
\begin{equation}
\min_{J,{\bf h}} {\rm KL}(P_{\mathcal{D}}({\bf x})|P({\bf x}|J,{\bf h})) = \min_{J,{\bf h}}\left\{ \sum_{\bf x} P_{\mathcal{D}}({\bf x}) \log \left(\frac{P_{\mathcal{D}}({\bf x})}{P({\bf x}|J,{\bf h})}\right)\right\}
\end{equation}
yields the maximum-likelihood estimation.
However, the computational time is excessive, because the method demands evaluation of the partition function depending on $J$ and ${\bf h}$.
Therefore, we require an effective technique to either approximate the partition function or avoid the computation of the partition function.

In the present study, we selected the latter technique.
One of the simplest methods to mitigate the computation of the log-likelihood function in Boltzmann machine learning is the pseudo-likelihood estimation \cite{Besag1975,Ekeberg2013}.
We change the cost function in the maximum-likelihood estimation, which has no terms in common with the partition function, as
\begin{equation}
\sum_{k=1}^D \log P({\bf x}^{(k)}|J,{\bf h} ) \approx \sum_{k=1}^D \log \prod_{i=1}^N P(x_i|J,{\bf h},{\bf x}^{(k)}_{/i}),
\end{equation}
where
\begin{equation}
P(x_i|J,{\bf h},{\bf x}_{/i}) = \frac{1}{Z_i(J,{\bf h}|{\bf x}_{/i})}\exp\left\{ \sum_{j \in  \partial i} J_{ij} x_i x_j + h_i x_i \right\}
\end{equation}
and
\begin{equation}
Z_i(J,{\bf h}|{\bf x}_{/i}) = \sum_{x_i}\exp\left( \sum_{j \in  \partial i} J_{ij} x_i x_j + h_i x_i \right) = 2\cosh\left( \sum_{j \in  \partial i} J_{ij} x_j + h_i\right).
\end{equation}
In the following, we deal with the minimization problem and take the negative of the approximated quantity as the cost function, that is
\begin{equation}
\mathcal{L}_{\rm PL}(J,{\bf h}) = - \sum_{k=1}^D \log \prod_{i=1}^N P(x_i|J,{\bf h},{\bf x}^{(k)}_{/i}).
\end{equation}
This appears to be a type of mean-field analysis, but the pseudo-likelihood estimation asymptotically (large amount of training data) coincides with the maximum-likelihood estimation.
This method is very simple and easy to implement, but requires an excessive amount of data.

Another technique for changing the cost function is the minimum probability flow \cite{Sohl-Dickstein2011}.
This method was inspired by relaxation dynamics, starting from the empirical distribution determined by the given data toward the distribution, using tentative parameters.
Relaxation dynamics are implemented by a master equation as
\begin{equation}
\frac{dP_t({\bf x})}{dt} = \sum_{\bf y}W({\bf x}|{\bf y})P_t({\bf y}),
\end{equation}
where $W({\bf x}|{\bf y})$ is the transition rate matrix.
We impose a one-spin flip at each update and detailed balance condition as
\begin{eqnarray}
W({\bf x}^{(l)}|{\bf x}^{(k)}) = \exp\left\{- \frac{1}{2}\left(E({\bf x}^{(l)}|J,{\bf h}) - E({\bf x}^{(k)}|J,{\bf h})\right)\right\} && {\rm for}~ \sum_{i=1}^Nx^{(k)}_ix^{(l)}_i = N-2,\label{TM}
\end{eqnarray}
where 
\begin{equation}
E({\bf x}|J,{\bf h}) = - \sum_{i=1}^N \sum_{j \in \partial i} J_{ij} x_i x_j - \sum_{i=1}^N h_ix_i.
\end{equation}
The choice of the transition matrix is very important in the following manipulation of the minimum probability flow.
The maximum likelihood estimation is computationally intractable due to the computation of the partition function.
We remove the dependence on the partition function by choosing the local update rule in the transition matrix as in Eq. (\ref{TM}).
For instance, the Metropolis method and heat-bath method can be applied to the minimum probability flow.
In the present study, we follow the original formulation of the minimum probability flow in the literature\cite{Sohl-Dickstein2011} for its symmetric form in computation as shown below.
If we tune the parameters adequately for the empirical distribution of the given data, the change from the initial distribution, namely the empirical distribution $P_{\mathcal{D}}({\bf x})$, is expected to be small; otherwise, it becomes large.
To capture this expectation, we then compute the following infinitesimal change of the KL divergence as
\begin{equation}
{\rm KL}(P_0({\bf x})|P_t({\bf x})) \approx {\rm KL}(P_0({\bf x})|P_0({\bf x})) + dt \frac{d}{dt}\left.{\rm KL}(P_0({\bf x})|P_t({\bf x}))\right\rvert_{t=0}.
\end{equation}
The combination of elementary algebra and the master equation leads up to the first order of $dt$ as
\begin{equation}
{\rm KL}(P_0({\bf x})|P_t({\bf x})) \approx \frac{dt}{D}\sum_{k=1}^D \sum_{l \notin \mathcal{D}|l \in \partial k}\exp\left\{ \frac{1}{2}\left(E({\bf x}^{(k)}|J,{\bf h})-E({\bf x}^{(l)}|J,{\bf h})\right)\right\}.\label{MPF}
\end{equation}
The true parameters are then estimated by minimization of this quantity.
This is the minimum probability flow method.
Notice that we do not require to manipulate the Markov chain Monte Carlo (MCMC) simulation, although the method is inspired by stochastic dynamics.
This is different from contrastive divergence, which requires computation by MCMC \cite{Welling2002}.
Once we impose the stochastic dynamics rule and the detailed balanced condition, we immediately compute the above quantity.
Thus, we utilize Eq. (\ref{MPF}) as the cost function to be minimized for estimating the parameters, instead of the log-likelihood function as in the maximum likelihood estimation; that is
\begin{eqnarray}
\mathcal{L}_{\rm MPF}(J,{\bf h}) = \frac{1}{D}\sum_{k=1}^D \sum_{l \notin \mathcal{D}}\exp\left\{ \frac{1}{2}\left(E({\bf x}^{(k)}|J,{\bf h})-E({\bf x}^{(l)}|J,{\bf h})\right)\right\},
\end{eqnarray}
where the summation over $l$ results in the case satisfying ${\sum_{i=1}^N x_i^{(k)}x_i^{(l)} = N-2}$.
The performance, estimation precision, and computational efficiency often exceed those of the pseudo-likelihood estimation for the same amount of data.
In the present study, we employ these methods to estimate the parameters; the following discussion can be straightforwardly applied to them.

Above all, we assume that parameters $J$ and ${\bf h}$ are assigned to all pairs and all components.
However, in order to elucidate the most relevant pair-wise interactions and biases from the given data, we employ an additional technique to prune less significant parameters.
A candidate is required to utilize the regularization of the $L_1$ norm \cite{Bishop2006}.
Let us then minimize the cost function $\mathcal{L}$ (=$\mathcal{L}_{\rm MPF}$ or $\mathcal{L}_{\rm PL}$) with $L_1$ norm as
\begin{equation}
\min_{J,{\bf h}}\left\{\lambda_J \sum_{(ij)}|J_{ij}| + \lambda_h \sum_{i=1}^N |h_i| + \mathcal{L}(J,{\bf h}) \right\}.
\end{equation}
The regularization technique was originally designed to obtain a unique estimation from underdetermined equations by imposing additional conditions.
Therefore, estimations that utilize regularization lead to stable solutions, even from small amounts of data.
As compensation, the entire cost function is not smooth, owing to the existence of the absolute value function.
The non-smoothness impedes the simple application of the gradient method, which identifies the minimal point of the cost function.
For the absolute value function, we may prepare several types of imitating functions.
However, this type of approximation does occasionally generate incorrect estimations, and reduces the convergence rate.
Instead of the original optimization problem with a non-smooth term, let us utilize a different function sharing the same optimal point below, that is the majorizer minimization.

\section{Majorizer minimization}
We briefly review majorizer minimization for convenience.
In general, we consider the optimization problem by minimizing a convex function $f$ with $N$-dimensional variables, which is assumed to be differentiable; its derivative $\nabla f({\bf x})$ is Lipschitz.
When the derivative is Lipschitz, there is a constant $L \ge 0$ for any ${\bf a}$ and ${\bf b}$
\begin{equation}
\sum_{k=1}^N\left( \left.\frac{\partial f}{\partial x_k}\right\rvert_{{\bf x}={\bf a}} - \left.\frac{\partial f}{\partial x_k}\right\rvert_{{\bf x}={\bf b}} \right)^2 \le L\sum_{k=1}^N\left(a_k - b_k \right)^2,
\end{equation}
where $L$ is termed as the Lipschitz constant and $a_k$ and $b_k$ are the $k$th component of $N$-dimensional vectors ${\bf a}$ and ${\bf b}$.
The majorizer of the function $f$ is then given by the following quadratic function
\begin{equation}
g({\bf x},{\bf v}) = f({\bf v}) + \sum_{k=1}^N \left.\frac{\partial f}{\partial x_k}\right\rvert_{{\bf x}={\bf v}}(x_k -v_k) + \frac{L}{2}\sum_{k=1}^N \left( x_k - v_k \right)^2.
\end{equation}
The majorizer always satisfies
\begin{equation}
f({\bf x}) \le g({\bf x},{\bf v}) \le f({\bf v}).
\end{equation}
Let us then consider the following optimization problem.
\begin{equation}
{\bf x}^{t+1} = \arg \min_{\bf x}\left\{ g({\bf x},{\bf x}^t) \right\}.
\end{equation}
The sequence of the optimal solutions $[{\bf x}^0,{\bf x}^1,\cdots,{\bf x}^T]$ satisfies
\begin{equation}
f({\bf x}^{t+1}) \le g({\bf x}^{t+1},{\bf x}^t) \le f({\bf x}^t)
\end{equation}
for $t=0,1,\cdots,T-1$.
This property of the majorizer gradually approaches the optimal solution of the original minimization problem.
This technique is referred to as the majorizer minimization approach, which is one of the gradient methods.
The convergence rate is known as $f({\bf x}^t) - f({\bf x}^*) = O(1/t)$, where the asterisk stands for the optimal solution.
When we utilize the regularization obtained with the $L_1$ norm, we solve the following optimization problem
\begin{equation}
{\bf x}^{t+1} = \arg \min_{\bf x}\left\{ g({\bf x},{\bf x}^t) + \lambda \left\| {\bf x}\right\|_1 \right\},
\end{equation}
where $\left\|{\bf x} \right\|_1 = \sum_{k=1}^N|x_k|$.
Because the majorizer is quadratic and the $L_1$ norm is separable, the optimal solution can be analytically obtained as
\begin{equation}
x_k^{t+1} = \eta_{\lambda/L}\left( x_k^{t} + \frac{1}{L} \left.\frac{\partial f}{\partial x_k}\right\rvert_{{\bf x}={\bf x}^t} \right),
\end{equation}
where
\begin{equation}
\eta_a(x)= {\rm sign}(x) (|x| - a).
\end{equation}
Therefore, solving alternative optimization problems is reduced to a simple substitution using the tentative solution ${\bf x}^t$.
The majorizer minimization method is broadly used in compressed sensing methods, which reconstruct original inputs from undersampled outputs.
In this problem, the original inputs should be sparse.
$L_1$-regularization enforces a sparse solution for the inference problem of the original signals.
Similarly, let us utilize the majorizer minimization method for estimation of the Boltzmann machine learning parameters.
Let us remark the role of the majorizer in short.
The majorizer modifies the original optimization problem into quadratic form.
The quadratic form separates the dependence on each component.
In other words, the many-body interaction system with the original function $f$ is changed into a one-body independent system consisting of the majorizer.
This is a type of mean-field analysis, which approximates the many-body interactions into an effective one-body description.
In statistical mechanics, the law of large numbers is imposed on the number of components $N$ to perform mean-field analysis and validation.
However, in this method, we do not require a large number of components; we only require the property of function $f$.
In this sense, it is a very generic yet powerful technique.

Let us apply the majorizer minimization approach to Boltzmann machine learning with $L_1$ regularization.
Because the pseudo-likelihood function and cost function in the minimum probability flow are differentiable and convex \cite{Sohl-Dickstein2011}, the majorizer minimization method can be applied.
The majorizer for Boltzmann machine learning is given as
\begin{eqnarray}\nonumber
G(J',{\bf h}';J,{\bf h}) &=& \mathcal{L}(J,{\bf h}) + \sum_{(ij)}\left.\frac{\partial \mathcal{L}(J,{\bf h})}{\partial J_{ij}}\right\rvert_{J,{\bf h}}\left(J'_{ij} - J_{ij} \right) + \frac{L_J}{2}\sum_{(ij)}\left(J'_{ij} - J_{ij}\right)^2
\\
&& \quad + \sum_{i}\left.\frac{\partial \mathcal{L}(J,{\bf h})}{\partial h_i}\right\rvert_{J,{\bf h}}\left(h'_{i} - h_{i} \right) + \frac{L_h}{2}\sum_{i}\left(h'_i - h_i \right)^2,
\end{eqnarray}
where $L_J$ and $L_h$ satisfy
\begin{eqnarray}
\sum_{(ij)}\left(\left.\frac{\partial \mathcal{L}(J,{\bf h})}{\partial J_{ij}}\right\rvert_{A,{\bf h}} - \left.\frac{\partial \mathcal{L}(J,{\bf h})}{\partial J_{ij}}\right\rvert_{B,{\bf h}} \right)^2 \le L_J \sum_{(ij)} \left(A_{ij} - B_{ij} \right)^2\\
\sum_{i}\left(\left.\frac{\partial \mathcal{L}(J,{\bf h})}{\partial h_{i}}\right\rvert_{J,{\bf a}} - \left.\frac{\partial \mathcal{L}(J,{\bf h})}{\partial h_{i}}\right\rvert_{J,{\bf b}} \right)^2 \le L_h \sum_{i} \left(a_{i} - b_{i} \right)^2.
\end{eqnarray}
Following the prescription of the majorizer minimization approach, let us iteratively solve the optimization problem
\begin{equation}
\left\{J^{t+1},{\bf h}^{t+1}\right\} = \arg \min_{J,{\bf h}}\left\{ \lambda_{J}\sum_{(ij)}|J_{ij}| + \lambda_h \sum_{i}|h_i| + G(J,{\bf h};J^{t},{\bf h}^t)\right\}.
\end{equation}
Because the dependence of $J$ and ${\bf h}$ on the majorizer is separate, we independently solve the optimization problem for each parameter as
\begin{eqnarray}
J_{ij}^{t+1} &=& \eta_{\lambda_J/L_J}\left( J_{ij}^{t} + \frac{1}{L_J}\left.\frac{\partial \mathcal{L}(J,{\bf h})}{\partial J_{ij}}\right\rvert_{J^t,{\bf h}^t}\right)\\
h_{i}^{t+1} &=& \eta_{\lambda_h/L_h}\left( h_{i}^{t} + \frac{1}{L_h}\left.\frac{\partial \mathcal{L}(J,{\bf h})}{\partial h_{i}}\right\rvert_{J^t,{\bf h}^t}\right).
\end{eqnarray}

The majorizer minimization method is a generic technique for reaching a minimum point by recursive manipulation, under the assumption that the cost function is convex and its derivative is Lipschitz.
These conditions are satisfied in the cost functions of the pseudo-likelihood function and minimum probability flow.
The derivatives of the pseudo-likelihood function yield
\begin{eqnarray}
- \frac{\partial \mathcal{L}_{\rm PL}(J,{\bf h})}{\partial J_{ij}} &=& \frac{1}{D}\sum_{k=1}^D x^{(k)}_ix^{(k)}_j - \frac{1}{D}\sum_{k=1}^D\sum_{i=1}^Nx_j^{(k)}\tanh\left( \sum_{j \in \partial i} J_{ij}x_j^{(k)} + h_i \right) \\
- \frac{\partial \mathcal{L}_{\rm PL}(J,{\bf h})}{\partial h_{i}} &=& \frac{1}{D}\sum_{k=1}^D x^{(k)}_i - \frac{1}{D}\sum_{k=1}^D\sum_{i=1}^N\tanh\left( \sum_{j \in \partial i} J_{ij}x_j^{(k)} + h_i \right).
\end{eqnarray}
In these cases, it is difficult to compute the Lipschitz constant.
We may use the backtracking technique, in which we gradually tune $L_J$ and $L_h$ by some rule such that
\begin{equation}
\mathcal{L}_{\rm PL}(J^{t+1},{\bf h}^{t+1}) \le G(J^{t+1},{\bf h}^{t+1}|J^t,{\bf h}^t).
\end{equation}

In addition, the case of the minimum probability flow is evaluated as
\begin{eqnarray}
\frac{\partial \mathcal{L}_{\rm MPF}(J,{\bf h})}{\partial J_{ij}} &=& \frac{1}{D}\sum_{k=1}^D \sum_{l \notin \mathcal{D} }\left( x^{(k)}_ix^{(k)}_j - x^{(l)}_ix^{(l)}_j \right)\exp\left\{\frac{1}{2}\left(E({\bf x}^{(k)}|J,{\bf h}) - E({\bf x}^{(l)}|J,{\bf h})\right)\right\} \label{MPF1}\\
\frac{\partial \mathcal{L}_{\rm MPF}(J,{\bf h})}{\partial h_{i}} &=& \frac{1}{D}\sum_{k=1}^D \sum_{l \notin \mathcal{D} }\left( x^{(k)}_i - x^{(l)}_i \right)\exp\left\{\frac{1}{2}\left(E({\bf x}^{(k)}|J,{\bf h}) - E({\bf x}^{(l)}|J,{\bf h})\right)\right\}.\label{MPF2}
\end{eqnarray}
These gradients are reduced for one-spin flips, using $\sum_{i=1}^N x_i^{(k)}x_i^{(l)} = N -2$
\begin{eqnarray}
\frac{\partial \mathcal{L}_{\rm MPF}(J,{\bf h})}{\partial J_{ij}} &=& \frac{2}{D}\sum_{k=1}^D\sum_{i =1}^N \sum_{j \in \partial i} x^{(k)}_ix^{(k)}_j  \exp\left\{\sum_{n \in \partial i}J_{ij}x^{(k)}_ix^{(k)}_n + h_i x^{(k)}_i\right) \\
\frac{\partial \mathcal{L}_{\rm MPF}(J,{\bf h})}{\partial h_{i}} &=& \frac{2}{D}\sum_{k=1}^D \sum_{i =1}^N x^{(k)}_i \exp\left\{\sum_{j \in \partial i}J_{ij}x^{(k)}_ix^{(k)}_j + h_i x^{(k)}_i\right).
\end{eqnarray}
where we assume that $i$th spin is flipped from the $k$th spin configuration (this is the $l$th configuration in the summation in Eqs. (\ref{MPF1}) and (\ref{MPF2})).
Similarly, we may use the backtracking technique such that $L_J$ and $L_h$ hold
\begin{equation}
\mathcal{L}_{\rm MPF}(J^{t+1},{\bf h}^{t+1}) \le G(J^{t+1},{\bf h}^{t+1}|J^t,{\bf h}^t).
\end{equation}
An acceleration technique is available for the majorizer minimization method \cite{Beck2009}.
We modify the update rule into
\begin{eqnarray}\nonumber
J_{ij}^{t+1} &=& \eta_{\lambda_J/L_J}\left( J_{ij}^{t} + \frac{1}{L_J}\left.\frac{\partial \mathcal{L}(J,{\bf h})}{\partial J_{ij}}\right\rvert_{J^t,{\bf h}^t}\right) + \left( \frac{\beta_{t} - 1}{\beta_{t+1}}\right)\left(\eta_{\lambda_J/L_J}\left( J_{ij}^{t} + \frac{1}{L_J}\left.\frac{\partial \mathcal{L}(J,{\bf h})}{\partial J_{ij}}\right\rvert_{J^t,{\bf h}^t}\right) - J_{ij}^t\right) \\
\\ \nonumber
h_i^{t+1} &=& \eta_{\lambda_h/L_h}\left( h_{i}^{t} + \frac{1}{L_h}\left.\frac{\partial \mathcal{L}(J,{\bf h})}{\partial h_{i}}\right\rvert_{J^t,{\bf h}^t}\right) + \left( \frac{\beta_{t} - 1}{\beta_{t+1}}\right)\left(\eta_{\lambda_h/L_h}\left( h_{i}^{t} + \frac{1}{L_h}\left.\frac{\partial \mathcal{L}(J,{\bf h})}{\partial h_{i}}\right\rvert_{J^t,{\bf h}^t}\right) - h_{i}^t\right),\\
\end{eqnarray}
where
\begin{equation}
\beta_{t+1} = \frac{1+\sqrt{1+4\beta_t^2}}{2}.
\end{equation}
The initial condition is $\beta_0 = 1$.
In this update rule, the convergence speed is improved as $\sum_{(ij)}\left( J^t_{ij} - J^*_{ij}\right)^2$ and $\sum_{i}\left( h^t_{i} - h^*_{i}\right)^2 \sim O(1/t^2)$, where the asterisk denotes the optimal solution.
\section{Numerical test}
We conducted several numerical experiments to test the estimation of sparse interactions.
The spin configurations were generated from the Markov chain Monte Carlo simulations.
The linear size $N_{L}=5$; that is, the entire spin $N = N_L^2 = 25$.
The number of interactions was $N^2=625$; the number of biases was $N=25$.
The true parameters for the biases were given by a Gaussian distribution with zero mean and unit variance.
In contrast, the true parameters for the interactions were restricted to (i)  the random sparse pairs (the non-zero interactions is restricted to $10\%$ of all pairs, namely $62$) and (ii) the nearest neighboring pairs on the square lattice (the number of non-zero interactions $100$).
We assumed that the interactions should be symmetric, namely $J_{ij} = J_{ji}$.
The values for the interactions used random variables that follow a Gaussian distribution with zero mean and unit variance.

\begin{figure}[tb]
\begin{center}
\includegraphics[width=140mm]{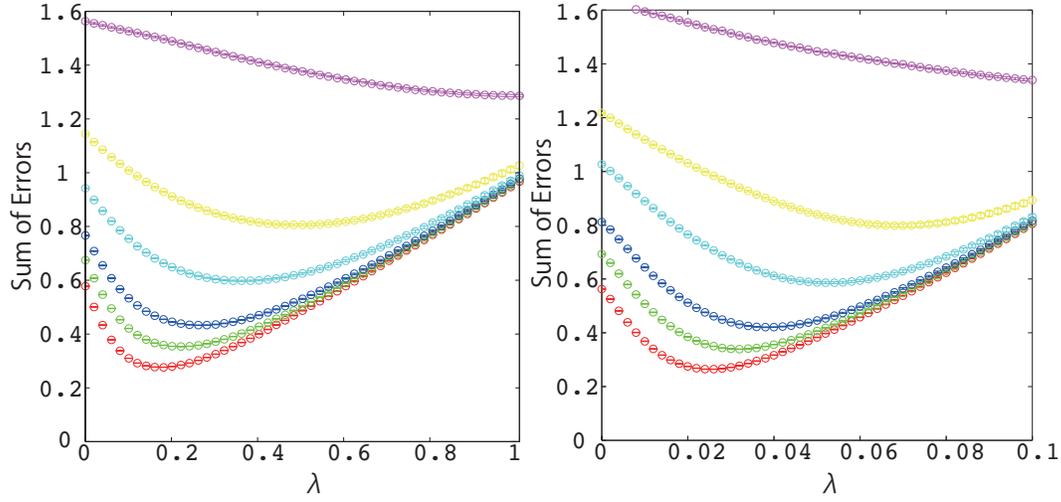}
\end{center}
\caption{{\protect\small (Color online)  Average performance of $L_1$-regularized inference in the case of random sparse interactions.
The horizontal axis denotes the amount of data. The vertical axis stands for the summation of the errors on estimations of $J$ and ${\bf h}$, ${\rm Err}_J+{\rm Err}_h$, which are defined as ${\rm Err}_J = \sqrt{\sum_{(ij)}\left(J_{ij} - J^{(\rm true)}_{ij}\right)^2/\sum_{(ij)}J_{ij}^2}$ and ${\rm Err}_h = \sqrt{\sum_{i}\left(h_i - h^{(\rm true)}_i\right)^2/\sum_{i}h_i^2}$.
The data amounts were $D=100$ (magenta), $D=500$ (yellow), $D=1000$ (cyan), $D=2000$ (red), $D=3000$ (green), and $D=5000$ (blue) from top to bottom.
}}
\label{fig1}
\end{figure}

The estimation had no prior knowledge of the structure of $J$ and ${\bf h}$.
In other words, the estimator did not know the lattice, and did not know that the non-zero interaction was restricted to specific pairs.
For each method, we estimated the parameters while changing $D$ as $D=100, 500$, $1000$, $2000$, $3000$, and $5000$.
The optimal selection of the coefficient $\lambda$ could not be known a priori.
We then tested several values of $\lambda$ for the estimations of the parameters.
In Fig. \ref{fig1}, we show the averaged performance over $100$ samples after $200$ iterations for the pseudo-likelihood estimation, and $50$ iterations for the minimum probability flow, for a case in which the pair-wise interactions were distributed randomly.
\begin{figure}[tb]
\begin{center}
\includegraphics[width=160mm]{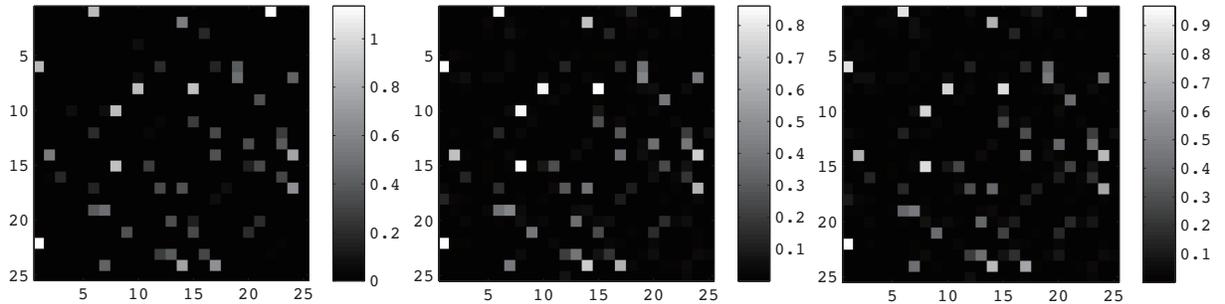}
\end{center}
\caption{{\protect\small (Color online) Profile (absolute value) of the pair-wise interactions in the random sparse case (one example).
The left panel shows the original configuration of the pair-wise interactions.
The center panel shows the results of the pseudo-likelihood estimation ($\lambda=0.2$); the right panel shows the results of the minimum probability flow ($\lambda=0.02$).
}}
\label{fig2}
\end{figure}
We note that the convergence speed of the minimum probability flow was significantly faster than the pseudo-likelihood estimation, although the precision of the convergent solutions was comparable.
The numbers of iterations used in both methods were sufficient to obtain the convergent estimations.
Both of the methods could estimate the correct values of the biases and interactions.
In Fig. \ref{fig2}, we show the profile of the estimated interactions for a single sample.
We confirmed that the estimation of the non-zero interactions had been achieved, although their absolute values tended to be smaller than the original values.
This is a characteristic property of the $L_1$ regularization.
We compared the pair-wise interactions and biases to the true parameters, as shown in Fig. \ref{fig3}.
We observe a fairly good performance for the nonzero components of the pair-wise interactions and biases.
The zeros of the pair-wise interactions are obtained as extremely small valued estimations.
We may set some thresholds to prune the irrelevant interactions in the estimation.
\begin{figure}[tb]
\begin{center}
\includegraphics[width=120mm]{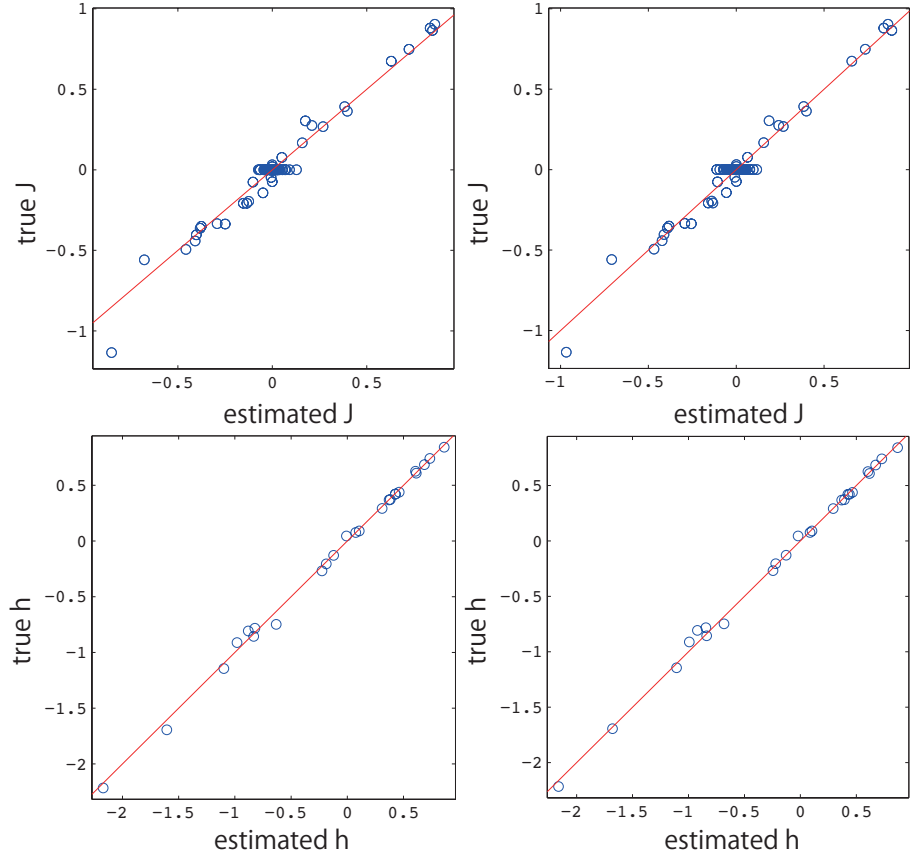}
\end{center}
\caption{{\protect\small (Color online) Comparison of the pair-wise interactions and biases to the true parameters in random sparse interactions (one example).
The vertical axis denotes the true parameters and the horizontal axis stands for the estimated values.
The upper left panel shows the results for $J_{ij}$ in the pseudo-likelihood estimation ($\lambda=0.2$) and the upper right one shows that of the minimum probability flow ($\lambda=0.02$).
The lower panels describe the results for $h_i$ by the pseudo-likelihood estimation (left) and the minimum probability flow (right).
The red lines have a unit slope as a guide to the eye.}}
\label{fig3}
\end{figure}
Figure \ref{fig4} shows the performance averaged over $100$ samples after $200$ iterations of the pseudo-likelihood estimation, and $50$ iterations of the minimum probability flow for a case in which pair-wise interactions were set on the square lattice.
An increase in $D$ improved the precision of the estimation in both methods.
Both methods could lead to precise estimations of the pair-wise interactions and biases.
The profile of the estimated interactions is shown in Fig. \ref{fig5}.
A comparison of the estimated interactions and biases with the true parameters is shown in Fig. \ref{fig6}.
We emphasize that the estimator did not have any prior knowledge of the structure of the interactions.
In this sense, we have succeeded in deriving the relevant structure of the pair-wise interactions from a type of microscopic degrees of freedom snapshot.
This indicates that the microscopic behavior observation characterized the generative model through the estimation, by use of $L_1$ regularization.
In addition, we truncated insignificant parameters with the aid of $L_1$ regularization.
In both cases of the random sparse interactions and the square lattice, we succeeded in reproducing the structure of the pair-wise interactions and estimating the magnitude of the interactions.
We emphasize that the gradient method with majorizer minimization method was replaced by the simple iterative substitution.
The technique we showed is expected to be applied to wide range of applications to seek the relevant interactions and biases generating the data.
In these numerical experiments, we demonstrate the case when we intend to apply our technique to the actual data.
Thus we prepare the specific pair-wise interactions a priori and generate the numerous data.
To further investigate the precision of our method, the hyperparameters $\lambda_J$ and $\lambda_h$ may be assumed to be distributed following the hyperprior distribution.
As shown above, we would find the least square error in the optimal hyperparameters, which correspond to the distributed ones. 

\begin{figure}[tb]
\begin{center}
\includegraphics[width=140mm]{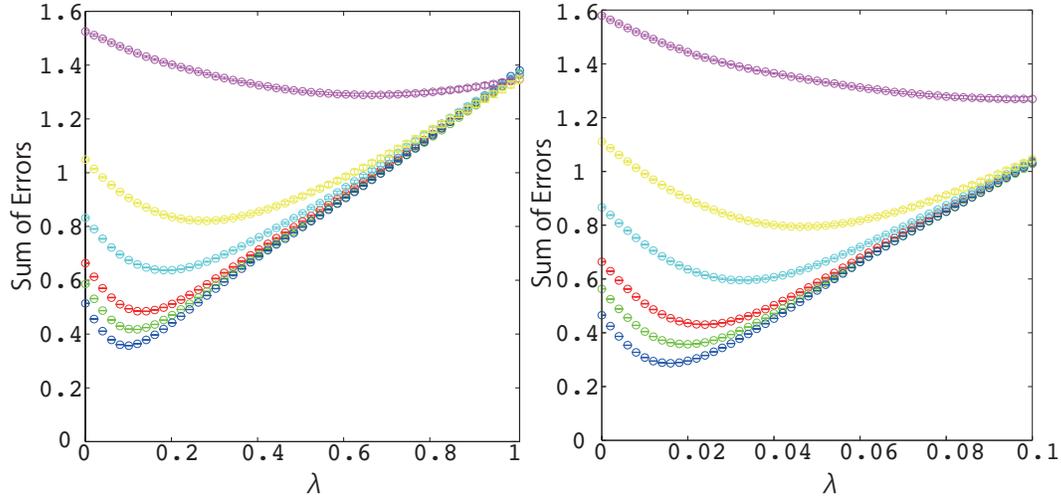}
\end{center}
\caption{{\protect\small (Color online)   Average performance of the $L_1$-regularized inference for a case in which a square lattice was used (one example).
The axes are the same as those in Fig. \ref{fig2}.
In this case, we further investigated the dependence on the amount of given data $D$.
The data amounts were $D=100$ (magenta), $D=500$ (yellow), $D=1000$ (cyan), $D=2000$ (red), $D=3000$ (green), and $D=5000$ (blue) from top to bottom.
}}
\label{fig4}
\end{figure}

\begin{figure}[tb]
\begin{center}
\includegraphics[width=160mm]{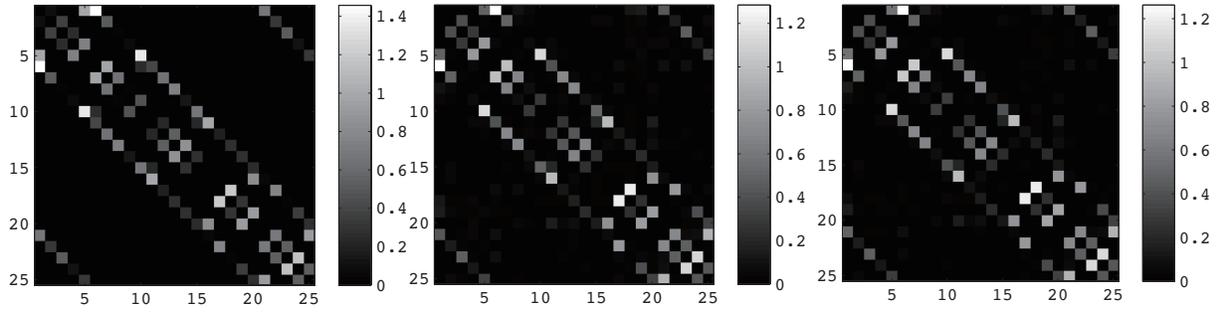}
\end{center}
\caption{{\protect\small Profile (absolute value) of the pair-wise interactions for a case in which a square lattice was used (one example).
The left panel shows the original configuration of the pair-wise interactions.
The center panel describes the estimation derived by the pseudo-likelihood estimation ($\lambda=0.1$) and the right panel shows the estimation derived by the minimum probability flow ($\lambda=0.018$).}}
\label{fig5}
\end{figure}

\begin{figure}[tb]
\begin{center}
\includegraphics[width=120mm]{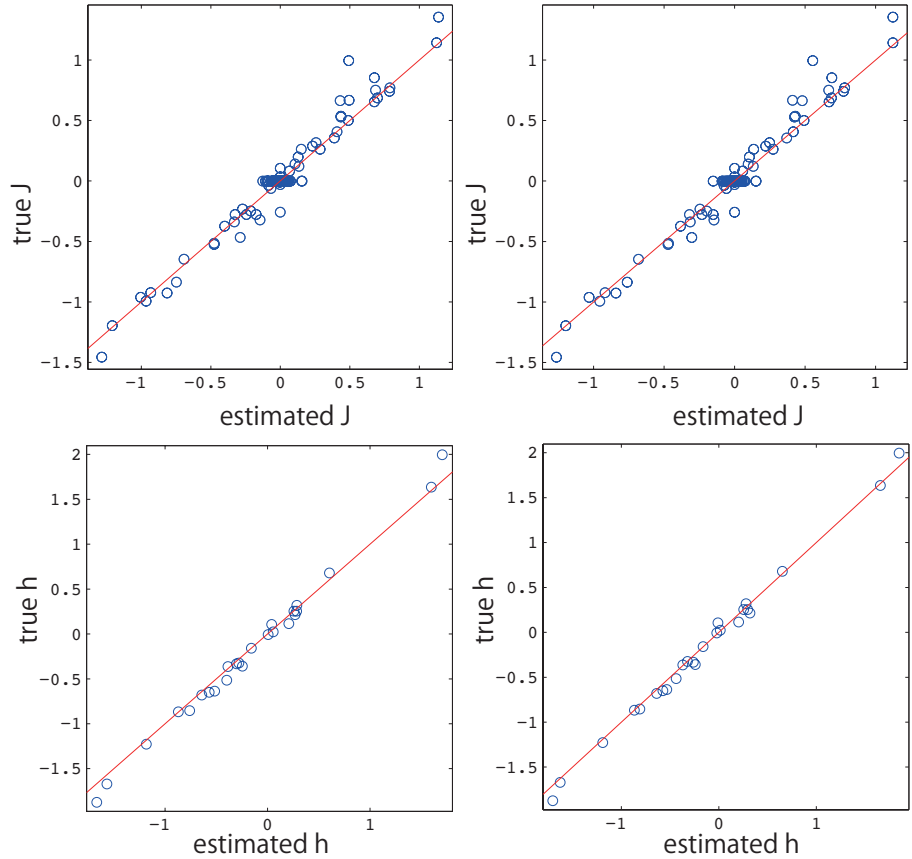}
\end{center}
\caption{{\protect\small Comparison of the pair-wise interactions and biases to the true parameters for a case in which a square lattice was used.
The symbols and axes are the same as those in Fig. \ref{fig3}}}
\label{fig6}
\end{figure}
\section{Summary}
In this study, we analyzed Boltzmann machine learning in terms of pseudo-likelihood estimation and minimum probability flow.
In order to elucidate the most relevant parameters generating the data, we sought a sparse solution in the present study.
This task was very important for determining the structure of the data while pruning irrelevant parameters.
$L_1$ regularization was beneficial in obtaining a sparse solution by solving a given cost function.
However, in general, the non-smoothness of the $L_1$ norm hampered the direct manipulation of the gradient method, which is intended to minimize the cost function.
This study featured the implementation of the majorizer minimization method into the Boltzmann machine learning technique.
The majorizer minimization method is a type of mean-field analysis, which enabled us to express a many-body interacting system in terms of an effective one-body independent system.

We tested our method to elucidate the randomly distributed interactions, and those between the adjacent spins on the square lattice, without any prior knowledge.
The performance of our method is fairly satisfactory, as expected.
Increasing the amount of given data improved the precision of the estimations and enhanced the efficacy of the $L_1$ regularization.
In present study, the cost functions are given by the pseudo likelihood function as well as the minimum probability flow.
The former one is generalized to the composite pseudo likelihood function inspired by the cluster variational method \cite{Yasuda2012proc}.
In this kind of generalization, the majorizer minimization is applicable.
In this sense, our scheme is very flexible.

Notice that our numerical experiments were assumed to be an extremely generic case, that is with in homogenous pair-wise interactions and biases.
One might intend to infer the homogeneous property from the given data.
The necessary number for precise estimations should then be extremely reduced.
The recent study improves precision of the Boltzmann machine learning with the comparable number of the data by aid of the Belief propagation to estimate the average and variance from the empirical data \cite{Yasuda2015}.
We anticipate that future studies will apply our present method to actual observed data, to elucidate the essential property from nature.
\section*{Acknowledgement}
The present work is performed by the financial support from the JST-CREST, MEXT KAKENHI Grants No. 251200008 and 24740263 and the Kayamori Foundation of Informational Science Advancement.
\bibliography{paper_ver2}

\begin{thebibliography}{10}

\bibitem{Ackley1985}
D.~H. Ackley, G.~E. Hinton, and T.~J. Sejnowski: Cognitive Science {\bfseries
  9} (1985) 147.

\bibitem{Hinton2006}
G.~E. Hinton, S.~Osindero, and Y.-W. Teh: Neural Comput. {\bfseries 18} (2006)
  1527.

\bibitem{Hinton2006sci}
G.~E. Hinton and R.~R. Salakhutdinov: Science {\bfseries 313} (2006) 504.

\bibitem{RG2014}
M.~Pankaj and J.~S. David:  {\bfseries stat.ML/1410.3831} (2014).

\bibitem{Ohzeki2015}
M.~Ohzeki: Journal of the Physical Society of Japan {\bfseries 84} (2015)
  034003.

\bibitem{Sessak2009}
V.~Sessak and R.~Monasson: Journal of Physics A: Mathematical and Theoretical
  {\bfseries 42} (2009) 055001.

\bibitem{Cocco2011}
S.~Cocco and R.~Monasson: Phys. Rev. Lett. {\bfseries 106} (2011) 090601.

\bibitem{Cocco2012}
S.~Cocco and R.~Monasson: Journal of Statistical Physics {\bfseries 147} (2012)
  252.

\bibitem{Ricci2012}
F.~Ricci-Tersenghi: Journal of Statistical Mechanics: Theory and Experiment
  {\bfseries 2012} (2012) P08015.

\bibitem{Yasuda2013}
M.~Yasuda and K.~Tanaka: Phys. Rev. E {\bfseries 87} (2013) 012134.

\bibitem{Raymond2013}
J.~Raymond and F.~Ricci-Tersenghi: Phys. Rev. E {\bfseries 87} (2013) 052111.

\bibitem{Ohzeki2013}
M.~Ohzeki: Journal of Physics: Conference Series {\bfseries 473} (2013) 012005.

\bibitem{Aurelien2014}
A.~Decelle and F.~Ricci-Tersenghi: Phys. Rev. Lett. {\bfseries 112} (2014)
  070603.

\bibitem{Yamanaka2015}
S.~Yamanaka, M.~Ohzeki, and A.~Decelle: Journal of the Physical Society of
  Japan {\bfseries 84} (2015) 024801.

\bibitem{Beck2009book}
A.~Beck and M.~Teboulle: in{\em Gradient-based algorithms with applications to
  signal-recovery problems}, ed. D.~P. Palomar and Y.~C. Eldar (Cambridge
  University Press, 2009), pp. 42--88.

\bibitem{Beck2009}
A.~Beck and M.~Teboulle: SIAM Journal on Imaging Sciences {\bfseries 2} (2009)
  183.

\bibitem{Bishop2006}
C.~Bishop: {\em Pattern recognition and machine learning} (Springer, New York,
  2006).

\bibitem{Besag1975}
J.~Besag: Journal of the Royal Statistical Society. Series D (The Statistician)
  {\bfseries 24} (1975) pp. 179.

\bibitem{Ekeberg2013}
M.~Ekeberg, C.~L\"ovkvist, Y.~Lan, M.~Weigt, and E.~Aurell: Phys. Rev. E
  {\bfseries 87} (2013) 012707.

\bibitem{Sohl-Dickstein2011}
J.~Sohl-Dickstein, P.~B. Battaglino, and M.~R. DeWeese: Phys. Rev. Lett.
  {\bfseries 107} (2011) 220601.

\bibitem{Welling2002}
M.~Welling and G.~Hinton: in{\em A New Learning Algorithm for Mean Field
  Boltzmann Machines}, ed. J.~Dorronsoro (Springer Berlin Heidelberg, 2002),
  Vol. 2415 of {\em Lecture Notes in Computer Science}, pp. 351--357.

\bibitem{Yasuda2012proc}
M.~Yasuda, S.~Kataoka, Y.~Waizumi, and K.~Tanaka: Pattern Recognition (ICPR),
  2012 21st International Conference on, Nov 2012, pp. 2234--2237.

\bibitem{Yasuda2015}
M.~Yasuda: Journal of the Physical Society of Japan {\bfseries 84} (2015)
  034001.

\end{thebibliography}
\end{document}